\documentclass[10pt,conference]{IEEEtran}
\IEEEoverridecommandlockouts

\usepackage{cite}
\usepackage{amsmath,amssymb,amsfonts}
\usepackage{algorithmic}
\usepackage{graphicx}
\usepackage{textcomp}
\usepackage{xcolor}
\usepackage{multirow}
\usepackage{booktabs}
\usepackage{url}
\def\BibTeX{{\rm B\kern-.05em{\sc i\kern-.025em b}\kern-.08em
    T\kern-.1667em\lower.7ex\hbox{E}\kern-.125emX}}
\begin{document}

\title{Multimodal Consistency-Guided Reference-Free Data Selection for ASR Accent Adaptation}

\author{
\IEEEauthorblockN{Ligong Lei, Wenwen Lu, Xudong Pang, Zaokere Kadeer, Aishan Wumaier}
\IEEEauthorblockA{\textit{School of Computer Science and Technology} \\
\textit{Xinjiang University}\\
Urumqi, Xinjiang, China}
}

\maketitle

\begin{abstract}
    Automatic speech recognition (ASR) systems often degrade on accented speech because acoustic-phonetic and prosodic shifts induce a mismatch to training data, making labeled accent adaptation costly. However, common pseudo-label selection heuristics are largely text-centric (e.g., perplexity (PPL) filtering) and can prefer fluent yet acoustically mismatched hypotheses, leading to error amplification when fine-tuning. To address this, we introduce a multimodal consistency-guided, reference-free data selection pipeline for ASR accent adaptation under a transductive, label-free protocol. The pipeline starts with a target-aware preselection step based on submodular mutual information to improve query relevance and reduce downstream computation. It then generates multiple pseudo-transcriptions per utterance via perturbation-based decoding and scores each hypothesis using two reference-free signals: speech--text alignment in a shared embedding space and predicted word error rate (WER). A simple percentile-based selection rule retains reliable pseudo-labels for fine-tuning while discarding noisy utterances. In an in-domain setting, selecting \(\sim\)1.5k utterances from a 30k pool achieves 10.91\% WER, close to 10.45\% obtained using 30k supervised labels. In a cross-domain setting with a mismatched candidate pool, consistency-filtered subsets avoid the degradation caused by unfiltered pseudo-labels under strong accent shift, and matched-hour experiments on a stronger ASR backbone further confirm gains over random sampling and recent selection baselines.
\end{abstract}


\begin{figure}[t]
  \centering
  \includegraphics[width=0.75\columnwidth]{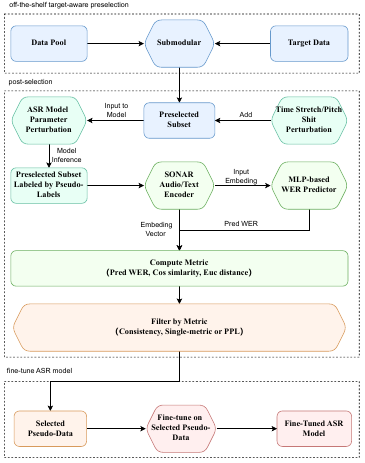}  
  \caption{Overall pipeline of our unsupervised data-selection framework.}
  \label{fig:pipeline}
\end{figure}

\section{Introduction}
\label{sec:intro}

Automatic Speech Recognition (ASR) enables many artificial intelligence applications, including voice assistants, captioning, call analytics, and accessibility tools. Recent progress in end-to-end modeling and self-supervised learning (SSL) pre-training has improved accuracy and robustness, yet ASR performance still drops substantially on accented speech. This failure is largely driven by accent-induced shifts in acoustic-phonetic realization and prosody that create a mismatch to training data. In practice, deploying high-performance accent-adapted ASR systems often requires substantial labeled data and compute, which increases cost and slows iteration.

Data selection provides a complementary direction for improving training efficiency on large-scale corpora by identifying and retaining the most informative samples. By removing redundant or noisy data, selection aims to reduce compute, time, and financial burden while maintaining, or even improving, generalization~\cite{Bartoldson2022ComputeEfficientDL}.

In accent adaptation, pseudo-labeling unlabeled target-domain speech is particularly attractive. However, selecting reliable pseudo-labels is non-trivial because accent mismatch primarily manifests in acoustics and prosody rather than in semantics. As a result, text-centric selection heuristics, for example filtering by language model perplexity, can over-prefer fluent hypotheses that remain acoustically inconsistent with the target accent, and thus fail to capture accent-induced acoustic discrepancies. 
In contrast, while acoustic realizations shift under accents, the underlying semantic content is typically preserved, which motivates using cross-modal consistency as a reference-free proxy for pseudo-label quality. 
Similar reference-free consistency cues have proven effective in other multimodal learning settings, where alignment between modalities provides a robust training signal even when direct supervision is limited or partially unreliable~\cite{shen2024imagpose,shen2024advancing} Prior work~\cite{shen2025imagdressing,shen2025imaggarment,shenlong}. on accent adaptation typically mitigates mismatch by fine-tuning a strong speaker-independent recognizer using a small amount of accent-specific labeled data~\cite{shor2019personalizing,sim2019investigation}, or by prioritizing sentences that are likely to trigger ASR errors for targeted recording and personalization~\cite{awasthi2021error}. These strategies, however, generally assume access to annotated target data. In this work, we instead study a pseudo-labeling setting where the goal is to identify reliable training utterances from a candidate pool without ground-truth transcripts.

We propose a selection pipeline that combines principled, query-centric coverage with reference-free quality estimation. We use Facility Location Mutual Information (FLMI)~\cite{kothawade2022prism} as an off-the-shelf, target-aware preselection step that improves query relevance and reduces downstream computation. Our main contribution is to show that multimodal consistency signals, specifically speech-to-text alignment in a shared embedding space using frozen SONAR~\cite{duquenne2023sonar}, together with reference-free WER prediction, provide more reliable pseudo-label quality estimation than text-only perplexity. This enables effective data selection for accent adaptation without ground-truth labels. Our contributions are threefold:
\begin{itemize}
\item We introduce a query-aware preselection stage based on FLMI to improve relevance and reduce the compute cost of downstream filtering.
\item We show that multimodal consistency, instantiated by frozen SONAR speech--text alignment and reference-free WER prediction, yields more reliable pseudo-label quality estimates than text-only perplexity in accent mismatch conditions.
\item We develop a simple and practical selection rule over multiple perturbed hypotheses per utterance, enabling label-free data selection for accent adaptation under a transductive protocol.
\end{itemize}
The code and data will be made available upon acceptance.

\section{Related Work}

\noindent\textbf{Data Selection.}
SSL-embedding-similarity-based selection measures self-supervised learning (SSL) embedding similarity to select data from a labeled source pool and boosts cross-domain ASR performance~\cite{lagos2024unsupervised}. However, this line of work focuses on labeled datasets and does not consider pseudo-label selection from unlabeled pools.
More practical pipelines first automatically transcribe unlabeled speech and then filter for valuable segments. Perplexity-contrast filtering retains domain-relevant pseudo-labeled utterances by contrasting the perplexity of a general language model against that of an in-domain language model~\cite{zheng2023unsupervised}.
A domain-classifier-based selection pipeline employs Whisper-X to obtain pseudo-transcriptions, uses TF-IDF and a support vector machine (SVM) for domain classification, and filters pseudo-labels based on word ratio and perplexity~\cite{rangappa2025speech}. While these methods are effective for domain adaptation, their text-centric heuristics often underrepresent acoustic-phonetic and prosodic factors central to accented speech, thus missing accent-induced acoustic discrepancies. DITTO uses submodular mutual information on Mel-frequency cepstral coefficient (MFCC) features to obtain fair accent-specific subsets, primarily selecting data from an acoustic perspective, but relies on human-annotated transcripts rather than pseudo-labels~\cite{kothawade2021ditto}. In a more recent study, WER-based pseudo-label filtering applies an SVM binary classifier with SSL features (0.5 threshold) and combines named entity recognition (NER) and character error rate (CER) for data selection~\cite{rangappa2025efficient}; although this approach considers the match between acoustic signals and pseudo-labels through WER prediction, the selection granularity remains coarse, relying primarily on binary accept/reject decisions rather than continuous, multi-factor quality scoring.

\noindent\textbf{Pseudo-label Quality Estimation.}
Reference-free WER estimation sees rapid progress in recent years. A reference-less ASR quality metric leverages known quality relationships between multiple compressed outputs of an ASR system, fine-tuned in a self-supervised manner using contrastive learning and a Siamese network architecture to pairwise rank ASR hypotheses~\cite{yuksel2023reference}. Meanwhile, FeWER~\cite{park2025fast} utilizes SSL features and a lightweight multilayer perceptron (MLP) to predict sentence-level WER; its system-agnostic variant further eliminates dependence on a specific decoder through a hypothesis generator~\cite{park2024automatic}. These advances lay a solid foundation for pseudo-label-based selection by providing reliable, reference-free quality signals. Our approach does not use WER prediction in isolation, but integrates it with cross-modal alignment scores to construct a unified consistency metric, which is then applied to data selection.

\noindent\textbf{Multimodal Representation.}
Sentence-level representation learning includes encoder-only, encoder-decoder, and teacher-student frameworks.
SONAR~\cite{duquenne2023sonar} employs a two-stage design: an encoder-decoder 
model first builds a multilingual text embedding space, after which speech encoders 
are trained in a teacher-student manner to align speech into the same space, enabling 
cross-lingual and cross-modal similarity search and translation mining. In our work, 
we utilize frozen SONAR speech/text encoders to obtain aligned sequence-level 
embeddings. This shared space enables measuring the semantic fidelity between speech and 
its pseudo-transcription, which in turn supports cross-modal, reference-free quality 
estimation within our selection pipeline.

Despite these advances, few works address \emph{label-free} pseudo-label selection for accent adaptation with quality signals that reflect speech--text consistency under accent-induced acoustic shifts. We propose a transductive, reference-free selection framework that leverages multimodal consistency signals, detailed next.

\section{Method}

Fig.~\ref{fig:pipeline} illustrates the data flow of our selection framework. Given a large candidate pool $\mathcal{C}$ and an unlabeled target query set $\mathcal{T}$, we (i) optionally apply a target-aware FLMI preselection (prior work) to obtain a reduced pool $\mathcal{P}\subseteq\mathcal{C}$; (ii) for each utterance $u\in\mathcal{P}$, decode a baseline hypothesis $H_0$ and a set of perturbed hypotheses $\{H_k\}$; (iii) score each hypothesis with reference-free signals---predicted WER and speech--text alignment in the shared SONAR embedding space; and (iv) apply the acceptance rule in Sec.~\ref{subsec:selection} to keep at most one hypothesis per utterance, yielding the final selected training subset $\mathcal{S}$ used for fine-tuning. Importantly, our contribution centers on steps (ii)--(iv), which can also be applied on a fixed candidate pool without FLMI.

\subsection{FLMI Preselection via Submodular Mutual Information}

We use FLMI as a target-aware preselection step following prior work~\cite{kothawade2022prism,kothawade2021ditto}. This stage is not our main contribution; rather, it provides query relevance and reduces the computational burden of the subsequent perturbation-and-consistency filtering. We briefly review the necessary definitions for completeness. A set function $f: 2^{\mathcal{U}} \rightarrow \mathbb{R}$ defined on a ground set $\mathcal{U}$ of $n$ data points is called \textit{submodular}~\cite{fujishige2005submodular} if it satisfies diminishing marginal returns: for all $\mathcal{X} \subseteq \mathcal{Y} \subseteq \mathcal{U}$ and $j \notin \mathcal{Y}$,
\begin{equation}
f(j \mid \mathcal{X}) = f(\mathcal{X} \cup \{j\}) - f(\mathcal{X}) \geq f(j \mid \mathcal{Y}).
\end{equation}
Submodularity ensures that a greedy algorithm achieves a bounded approximation factor when maximizing such a function~\cite{nemhauser1978analysis}.

To measure the similarity between a selected subset $\mathcal{S}$ and a target (or query) set $\mathcal{T}$, we use the Submodular Mutual Information (SMI)~\cite{gupta2020, iyer2019memoization}:
\begin{equation}
I_f(\mathcal{S}; \mathcal{T}) = f(\mathcal{S}) + f(\mathcal{T}) - f(\mathcal{S} \cup \mathcal{T}).
\end{equation}

In our work, we use the Facility Location Mutual Information (FLMI) function~\cite{kothawade2022prism}:
\begin{equation}
I_f(\mathcal{S}; \mathcal{T}) = \sum_{i \in \mathcal{T}} \max_{j \in \mathcal{S}} s_{ij} + \sum_{i \in \mathcal{S}} \max_{j \in \mathcal{T}} s_{ij},
\end{equation}
where $s_{ij}$ denotes the cosine similarity between the MFCC feature vectors of utterances $i$ and $j$. FLMI jointly captures both \textit{query-relevance} and \textit{representativeness}~\cite{kothawade2022prism}. In our transductive, label-free selection setting, the unlabeled target audio serves as the query set.

We follow DITTO~\cite{kothawade2021ditto} by extracting utterance-level 39-d mean MFCC features, precomputing all candidate-query cosine similarities, and maximizing the FLMI function $I_f(\mathcal{S}; \mathcal{T})$ using a LazyGreedy and out-of-core exact greedy algorithm combination.

Let $\mathcal{C}$ denote the source candidate pool and $\mathcal{T}$ the unlabeled target query set. Running FLMI with $\mathcal{T}$ as the query yields a target-aware preselected subset $\mathcal{P} \subseteq \mathcal{C}$. Unless stated otherwise, we set the FLMI preselection budget to $|\mathcal{P}|=30$k utterances. All subsequent selection, perturbation-based scoring, and fine-tuning operate on $\mathcal{P}$ or its subsets. We refer to the collection of all baseline and perturbed hypotheses constructed from $\mathcal{P}$ as the perturbation pool $\mathcal{Q}_{\mathcal{P}}$.

\subsection{Perturbation-based Inference}
For each utterance in the preselected subset $\mathcal{P}$, we decode a baseline hypothesis $H_0$ and a set of perturbed hypotheses $\{H_k\}_{k=1}^{K}$ generated by (i) model-parameter perturbation and (ii) input perturbation, forming the perturbation pool $\mathcal{Q}_{\mathcal{P}}$. 
Model perturbation injects zero-mean Gaussian noise into each parameter tensor $W$ with scale $\alpha\,\mathrm{std}(W)$, where $\alpha\in\{0.01,0.02,0.03\}$; the baseline corresponds to $\alpha=0.00$ (no parameter noise). 
Input perturbation (16 kHz) uses six variants: pitch-only shifts $\{-2,+1,+2\}$ semitones; and stretch--pitch pairs $(\text{atempo},\,\text{pitch})\in\{(0.90,-1),\,(0.95,-2),\,(0.95,-1)\}$. 
Overall, we generate 28 hypotheses per utterance before de-duplication: 1 baseline ($\alpha{=}0.00$ with no input perturbation), 6 input-perturbed hypotheses at $\alpha{=}0.00$, and $3\times7=21$ model-perturbed hypotheses over $\alpha\in\{0.01,0.02,0.03\}$ applied to the original plus six input variants. We then deduplicate identical transcriptions, so $K$ denotes the number of unique perturbed hypotheses kept (with $K \le 27$). 
All hypotheses in $\mathcal{Q}_{\mathcal{P}}$ are subsequently scored by the multimodal, reference-free consistency metrics described in Sec.~\ref{subsec:embedding} and filtered by the acceptance rule in Sec.~\ref{subsec:selection} to yield the final selected training subset.

\textit{Choice of perturbation types and scales.} We select perturbation types and scales based on an exploratory study on Common Voice using Whisper-base~\cite{radford2023robust} to obtain diverse decoding variants with non-trivial improvement rates (details in Sec.~\ref{subsec:asr_ft}).

\subsection{Sequence-level Embedding and WER Prediction}\label{subsec:embedding}

To obtain aligned speech--text representations, we use frozen pre-trained SONAR~\cite{duquenne2023sonar} encoders, which map audio and text inputs into a shared 1024-d language-agnostic embedding space.

We then train a lightweight reference-free WER predictor by concatenating the speech and text embeddings (2048-d input) and processing them through a 3-layer MLP (2048$\to$600$\to$32$\to$1) with BatchNorm--Linear--ReLU--Dropout blocks. A sigmoid output with learnable temperature $\beta$ constrains predictions to [0.01, 0.99].

Given the perturbation pool $\mathcal{Q}_{\mathcal{P}}$ for each utterance, we compute SONAR embeddings for the speech and for every hypothesis $H \in \mathcal{Q}_{\mathcal{P}}$. For each perturbed hypothesis $H_k$ we obtain a bounded WER estimate 
$\hat{w}_{k}=f_{\theta}\bigl([\mathbf{h}^{\text{sp}}\Vert\mathbf{h}^{\text{txt}}(H_{k})]\bigr)$. 
In parallel, we compute two alignment scores between the speech embedding $\mathbf{h}^{\text{sp}}$ and the text embedding $\mathbf{h}^{\text{txt}}(H_{k})$: cosine similarity $c_{k}$ and Euclidean distance $d_{k}$. We also compute the same alignment scores for the baseline $H_0$, yielding the baseline values used in Sec.~\ref{subsec:selection}. The resulting per-hypothesis quality vectors $\{\hat{w}_k, c_k, d_k\}$ annotate $\mathcal{Q}_{\mathcal{P}}$ for selection:

We compute cosine similarity as:
\begin{equation}
c_{k} = \frac{\langle \mathbf{h}^{\text{sp}}, \mathbf{h}^{\text{txt}}(H_{k}) \rangle}
              {\|\mathbf{h}^{\text{sp}}\|\,\|\mathbf{h}^{\text{txt}}(H_{k})\|},
\end{equation}
and Euclidean distance as:
\begin{equation}
d_{k} = \|\mathbf{h}^{\text{sp}}-\mathbf{h}^{\text{txt}}(H_{k})\|_{2}.
\end{equation}
These three quantities $\{\hat{w}, c, d\}$ constitute a quality vector that is passed to the selection procedure detailed in Sec.~\ref{subsec:selection}.

\subsection{Selection Strategy}\label{subsec:selection}

For each utterance $u\in\mathcal{P}$, we compare its perturbed hypotheses $\{H_k\}$ against the baseline $H_0$. To do so, we define per-metric improvements. Let $v_k$ and $v_0$ denote the metric values for $H_k$ and $H_0$, respectively, where $v\in\{\hat{w},c,d\}$. The improvement for metric $v$ is:

\begin{equation}
\Delta^{(v)}_k =
\begin{cases}
  v_0 - v_k, & v \in \{\hat{w}, d\} \\
  v_k - v_0, & v \in \{c\}
\end{cases}.
\label{eq:delta-def}
\end{equation}

By aggregating only positive improvements over the perturbation pool $\mathcal{Q}_{\mathcal{P}}$ induced by the same FLMI pre-selected subset $\mathcal{P}$, we estimate empirical distributions and compute percentile thresholds $\tau_{v}^{p}$. The thresholds are calculated per metric, ensuring that the selection is adaptive to the score distribution of the current candidate pool.

At runtime, a hypothesis is accepted if its quality metrics exceed empirical percentile thresholds. We experiment with multiple selection rules: (i) \emph{single-metric rules} using only one of $\{\Delta^{(\hat{w})}_k, \Delta^{(c)}_k, \Delta^{(d)}_k\}$, and (ii) a \emph{conjunction rule} requiring both predicted-WER improvement and at least one alignment improvement:
\begin{equation}
  \Delta^{(\hat{w})}_k \geq \tau_{\hat{w}}^{p}
  \;\land\;
  \bigl(
  \Delta^{(c)}_k \geq \tau_{c}^{p}
  \;\lor\;
  \Delta^{(d)}_k \geq \tau_{d}^{p}
  \bigr).
  \label{eq:accept-rule}
  \end{equation}
If multiple perturbations satisfy the rule for a given utterance, we keep the candidate with 
the largest predicted-WER improvement, i.e., the largest $\Delta^{(\hat{w})}_k$.

Applying the acceptance rule within $\mathcal{Q}_{\mathcal{P}}(u)$ for every $u\in\mathcal{P}$ yields at most one accepted hypothesis per utterance. The accepted set forms the final selected training subset $\mathcal{S} \subseteq \mathcal{P}$ with transcripts given by the retained hypotheses; utterances with no accepted hypothesis are excluded.

\section{Experiments}

\subsection{Experimental Setup}

We adopt a matched-budget comparison protocol: all methods are trained under the same fine-tuning budget to isolate the effect of selection and pseudo-label quality. The CRDNN/Wav2Vec~2.0 (W2V2) experiments (Table~\ref{tab:wer_two_datasets}) use a fixed 2-epoch budget; the Paraformer experiments (Sec.~\ref{subsec:paraformer}) provide a full budget sweep across hours.

\noindent\textbf{Baselines.} We group baselines by their filtering signal:

\noindent\textit{Training references:}
\textit{Pretrained} (no fine-tuning); 
\textit{All-Ref} (supervised oracle with ground-truth labels); 
\textit{All-Pseudo} (unfiltered pseudo-labels, null hypothesis).

\noindent\textit{PPL-based selection:}
\textit{PPL-P90} (text-only perplexity at matched percentile $p{=}90$);
\textit{PPL-SizeMatched} (same subset size as our method).

\noindent\textit{Multimodal selection (ours):}
\textit{Conf-P} (conjunction rule); 
\textit{Pred/Cos/Euc-Only} (single-metric variants);
\textit{Stable-Base-P} (baseline quality selection, Sec.~\ref{subsec:paraformer}).

\noindent\textit{Non-PPL baselines (Sec.~\ref{subsec:paraformer}):}
\textit{CER-Consistency} (model-agreement across 3 ASRs);
\textit{SSL-WER-Classifier} (SVM on SSL features);
\textit{Random} (random sampling at matched hours).

This grouping guides our analysis: Sec.~\ref{subsec:asr_ft} compares multimodal signals against PPL-based filtering under a fixed 2-epoch budget; Sec.~\ref{subsec:paraformer} provides a budget-sweep analysis including comparisons to both PPL and non-PPL baselines.

\textbf{Perturbation selection.} We select perturbation types and scales via an exploratory sweep on Common Voice using Whisper-base. Table~\ref{tab:perturbation_selection} summarizes the explored and selected parameters for each perturbation type. For each perturbation configuration, we decode both the baseline hypothesis (no perturbation) and the perturbed hypothesis, compute WER against references, and report (i) the \emph{improvement rate}, i.e., the fraction of utterances whose WER decreases relative to the baseline, and (ii) the average WER reduction over the improved utterances. We retain configurations with improvement rate $\geq$20\% and select the final six input variants by ranking configurations by improvement rate. Specifically, the selected input variants are the top six by improvement rate among the retained configurations: three pitch-only shifts $\{-2,+1,+2\}$ semitones, and three stretch--pitch pairs $\{(0.90,-1), (0.95,-2), (0.95,-1)\}$.

\begin{table}[t]
  \centering
  \renewcommand{\arraystretch}{1.05}
  \caption{Perturbation search space and selected parameters.}
  \label{tab:perturbation_selection}
  \begin{tabular}{l l l}
    \hline
    \textbf{Type} & \textbf{Explored} & \textbf{Selected} \\
    \hline
    Model noise $\alpha$ & \{0.01, 0.02, 0.03, 0.04, 0.05\} & \{0.01, 0.02, 0.03\} \\
    Pitch shift (st) & \{$-$2, $-$1, +1, +2\} & \{$-$2, +1, +2\} \\
    Time stretch & \{0.90, 0.95, 1.05, 1.10\} & \{0.90, 0.95\} \\
    Additive noise & \{0.005, 0.01, 0.02, 0.03\} & --- (excluded) \\
    \hline
  \end{tabular}
\end{table}

\subsection{Dataset}

We use three English corpora.
\textbf{IndicTTS English subset}~\cite{indictts2023}\footnote{\url{https://hf-mirror.com/SPRINGLab}}.
Originally designed for multilingual text-to-speech (TTS), the IndicTTS corpus includes professionally recorded and transcribed English read speech by both male and female speakers. Its English subset contains approximately 150k utterances covering 13 Indian accents. We split this subset 80\%/20\% to form a candidate pool for data selection and a held-out target split for evaluation, respectively. An auxiliary labeled validation set of 6,000 utterances is sampled from the candidate pool for early stopping. In this IndicTTS setting, we set the candidate pool $\mathcal{C}$ to the 80\% split and treat the held-out 20\% target audio as unlabeled for transductive querying.
\textbf{CSTR VCTK Corpus}~\cite{VCTKv092}.
This corpus contains recordings from 110 English speakers with diverse accents, each reading about 400 sentences. It serves as an additional candidate pool.
\textbf{L2-ARCTIC corpus}~\cite{zhao2018l2arctic}.
L2-ARCTIC contains 24 non-native English speakers (balanced gender and L1s, i.e., native languages) with 26,867 utterances and 27.1 hours in total; L1s include Arabic, Chinese, Hindi, Korean, Spanish, and Vietnamese. We resample all audio to 16 kHz.
For the L2-ARCTIC setting, we use VCTK as the candidate pool $\mathcal{C}$. We adopt a speaker-disjoint protocol on L2-ARCTIC: for each L1 accent, we reserve one speaker for dev and one speaker for test. We treat the test audio as unlabeled for transductive querying (with transcripts withheld) and report WER on this held-out test split; the dev split is used for early stopping. This dev/test split is created once and shared across both the CRDNN/W2V2 and Paraformer experiments.

We adopt a transductive, label-free selection protocol. This means the target audio is treated as unlabeled and used as queries for FLMI-based pre-selection from the candidate pools. The percentile thresholds $\tau^p$ used in our acceptance rule are computed \emph{only} from the pre-selected perturbation pool $\mathcal{Q}_{\mathcal{P}}$ constructed on $\mathcal{P}$ (Sec.~\ref{subsec:selection}); thus, target audio affects $\tau^p$ only indirectly through the target-aware pre-selected pool $\mathcal{P}$. Ground-truth labels are never used for selection, and the proposed method is fine-tuned using pseudo-labels (with labeled dev used only for early stopping). For the supervised oracle baseline (All-Ref), ground-truth labels are used only for the pre-selected subset from the candidate pool; labels are otherwise used only to compute WER for evaluation.

\subsection{Models and Fine-tuning}

We fine-tune three pretrained ASR backbones. We use SpeechBrain~\cite{ravanelli2021speechbrain} for CRDNN and W2V2 fine-tuning. Table~\ref{tab:training_config} summarizes training configurations.

\noindent(i) \textbf{CRDNN-RNNLM} (SpeechBrain/LibriSpeech~\cite{panayotov2015librispeech}): 2 conv + 4 BiLSTM + 2 FC encoder with GRU decoder over 1k BPE. Fine-tuned with hybrid CTC/sequence loss (label smoothing), AdamW + New-Bob scheduling; inference uses beam search with auxiliary RNNLM.

\noindent(ii) \textbf{Wav2Vec 2.0 + DNN} (SpeechBrain/Common Voice~\cite{ardila2020common}): frozen wav2vec2 2.0~\cite{baevski2020wav2vec} frontend, 2 FC + GRU with dual CTC/attention heads. Fine-tuned with Adam + New-Bob; beam search at inference (no external LM).

\noindent(iii) \textbf{Paraformer}~\cite{gao2022paraformer} (FunASR~\cite{gao2023funasr}): non-autoregressive model with Conformer encoder and bidirectional Transformer decoder. Fine-tuned from \texttt{paraformer-en} checkpoint with Adam + warmup scheduling.

\begin{table}[t]
  \centering
  \renewcommand{\arraystretch}{1.05}
  \caption{Training configurations for fine-tuning.}
  \label{tab:training_config}
  \begin{tabular}{l c c c c}
    \hline
    \textbf{Model} & \textbf{Epochs} & \textbf{LR} & \textbf{Batch} & \textbf{Optimizer} \\
    \hline
    CRDNN~\cite{ravanelli2021speechbrain} & 2 & \texttt{3e-5} & 16 & AdamW \\
    W2V2~\cite{ravanelli2021speechbrain} & 2 & \texttt{3e-5} & 16 & Adam \\
    Paraformer~\cite{gao2022paraformer} & 6 & \texttt{3e-5} & 64 & Adam \\
    \hline
  \end{tabular}
\end{table}

\subsection{Metric Computation Results}

The WER predictor was trained on Common Voice~\cite{ardila2020common} utterances transcribed by the Whisper-base~\cite{radford2023robust} model using uncertainty inference, 
with reference WER values clamped to the range [0, 1]. 
We used the AdamW optimizer with cosine annealing, MSE loss, Xavier initialization, and a dropout rate of 0.3. 
We employed early stopping, halting training at 70 epochs. Although trained out of domain, it shows high correlation and low error on our target datasets (Table~\ref{tab:metrics}), supporting its use as the core metric alongside cross-modal alignment for consistency filtering. 
\setlength{\tabcolsep}{6pt}\renewcommand{\arraystretch}{1.0}
\begin{table}[t]
  \centering
  \renewcommand{\arraystretch}{1.1}
  \caption{WER-predictor performance on preselected perturbed hypothesis sets.}
  \label{tab:metrics}
  \begin{tabular}{llcccc}
    \hline
    \textbf{Dataset} & \textbf{Model} & \textbf{Pearson} & \textbf{Spearman} & \textbf{MAE} & \textbf{RMSE}\\ \hline
    \multirow{2}{*}{IndicTTS~\cite{indictts2023}} & CRDNN & 0.86 & 0.77 & 0.09 & 0.13 \\
                           & W2V2 & 0.84 & 0.84 & 0.12 & 0.17 \\ \cline{2-6}
    \multirow{2}{*}{VCTK~\cite{VCTKv092}} & CRDNN & 0.85 & 0.85 & 0.14 & 0.21 \\
                           & W2V2 & 0.94 & 0.81 & 0.09 & 0.13 \\ \hline
  \end{tabular}
\end{table}

\begin{table*}[t]
  \centering
  \renewcommand{\arraystretch}{1.0}
  \caption{WER (\%) under a matched fine-tuning budget (2 epochs) for CRDNN and W2V2 in an in-domain and a cross-domain setting. Selection is transductive and label-free; All-Ref is a supervised oracle. Unless noted, we use $p{=}90$ (see Sec.~\ref{subsec:selection} and Sec.~\ref{subsec:asr_ft} for protocol details).}
  \label{tab:wer_two_datasets}
  \begin{tabular}{lc cccc cccc}
  \toprule
  \multirow{2}{*}{\textbf{Stage / Strategy}} &
  \multirow{2}{*}{\textbf{Method}} &
  \multicolumn{4}{c}{\textbf{IndicTTS}} &
  \multicolumn{4}{c}{\textbf{L2-ARCTIC}} \\
  \cmidrule(lr){3-6}\cmidrule(lr){7-10}
  & & \textbf{CRDNN} & \textbf{Utts} & \textbf{W2V2} & \textbf{Utts} & \textbf{CRDNN} & \textbf{Utts} & \textbf{W2V2} & \textbf{Utts} \\
  \midrule
  \multirow{1}{*}{Pre-Train}
  & Pretrained (no FT) & 14.90 & - & 12.03 & - & 22.97 & - & 15.31 & - \\
  \hline
  \multirow{2}{*}{Baseline}
  & All-Ref & 9.10 & 30k & 10.45 & 30k & 25.96 & 30k & 16.42 & 30k \\
  & All-Pseudo & 30.11 & 30k & 12.52 & 30k & 32.73 & 30k & 17.24 & 30k \\
  \hline
  \multirow{4}{*}{Consistency}
  & Conf-P95 & 12.72 & 911 & 11.23 & 282 & \textbf{22.00} & 687 & 15.32 & 331 \\
  & Conf-P90 & \textbf{12.70} & 1751 & 11.27 & 712 & 22.22 & 1225 & 15.32 & 485 \\
  & Conf-P80 & 13.08 & 3504 & \textbf{10.91} & 1519 & 22.13 & 2318 & \textbf{15.19} & 882 \\
  & Conf-P70 & 15.13 & 5190 & 11.11 & 2370 & 22.50 & 3505 & 15.21 & 1268 \\
  \hline
  \multirow{4}{*}{Single-Metric}
  & Conf-P90 & \textbf{12.70} & 1751 & 11.27 & 712 & 22.22 & 1225 & 15.32 & 485 \\
  & Pred-Only & 13.25 & 2317 & 11.30 & 1115 & 22.05 & 1525 & \textbf{15.19} & 571 \\
  & Cos-Only & 12.87 & 2876 & \textbf{10.90}  & 1014 & \textbf{20.03} & 1898 & 15.33 & 638 \\
  & Euc-Only & 12.83 & 3165 & \textbf{10.90}  & 1088 & 21.93 & 1949 & 15.26 & 614 \\
  \hline
  \multirow{3}{*}{Contrast}
  & Conf-P90 & \textbf{12.70} & 1751 & \textbf{11.27} & 712 & \textbf{22.22} & 1225 & 15.32 & 485 \\
  & PPL-P90 (p-matched) & 15.44 & 1499 & 11.44 & 782 & 25.18 & 1269 & \textbf{15.31} & 494 \\
  & PPL-SizeMatched & 15.25 & 1751 & 11.29 & 712 & 24.71 & 1225 & 15.32 & 485 \\
  \bottomrule
  \end{tabular}
  \end{table*}

\subsection{Fixed-epoch Evaluation}\label{subsec:asr_ft}

Experiments cover two models and two datasets (Table~\ref{tab:wer_two_datasets}).

\noindent\textbf{Baselines.} All-Pseudo degrades substantially on both datasets compared to Pretrained, especially for CRDNN, indicating that unfiltered pseudo-labels introduce harmful noise. All-Ref shows that in-domain labeled data (IndicTTS) provides positive gains, while out-of-domain labeled data (L2-ARCTIC with VCTK source) even hurts performance compared to zero-shot, suggesting negative transfer under domain mismatch.

\textbf{Key observations.}
(i) Effectiveness and model dependence (IndicTTS). Consistency-filtered subsets (Conf-P$^p$)
yield substantial gains over unfiltered pseudo-label training, especially for the weaker CRDNN
backbone, which is more susceptible to pseudo-label noise. With W2V2, using ~1.5k selected utterances from a 30k pool attains 10.91\% WER, close to 10.45\% from 30k supervised labels. Across the pre-specified percentile sweep, WERs vary only slightly (10.91--11.27\%), suggesting perturbation-consistent utterances provide high-quality training signals in this in-domain setting.
(ii) Behavior under domain shift (L2-ARCTIC). Under strong accent shift, consistency-filtered subsets avoid the degradation observed with unfiltered pseudo-labels: for CRDNN, Conf-P95 reaches 22.00\% while Cos/Euc achieve 20.03\%/21.93\%; for W2V2, results cluster tightly around the zero-shot baseline (15.19--15.33\%). Given the candidate domain (VCTK) is acoustically simpler than L2-ARCTIC, extracting helpful pseudo-labels is non-trivial, yet filtered subsets provide measurable benefit, highlighting stability rather than degradation.
(iii) Multimodal signals vs.\ text-only filtering. Multimodal variants generally outperform text-only PPL filtering for CRDNN and remain competitive for W2V2 (Table~\ref{tab:wer_two_datasets}, Contrast rows), supporting that speech--text alignment and predicted WER capture quality information complementary to lexical perplexity.

\textit{Note on ablations.} Fixing the same percentile threshold $p$ can yield different selected subset sizes across rules (Table~\ref{tab:wer_two_datasets}); we therefore interpret single-metric comparisons primarily as diagnostics of signal informativeness. Among single-metric variants, alignment-based signals (Cos-Only, Euc-Only) often perform competitively with the conjunction rule, suggesting that the primary value lies in the multimodal signals rather than the specific combination logic. This observation also motivates our use of matched audio-hours (rather than matched utterance counts) in Sec.~\ref{subsec:paraformer} for more equitable budget comparisons.

In summary, under a fixed 2-epoch fine-tuning budget, selection benefits are modulated by model capacity and domain difficulty: larger absolute gains arise in-domain (especially for CRDNN), while under strong shift the effect is primarily to mitigate error amplification without harming zero-shot performance. Across all settings, multimodal signals generally outperform text-only PPL, validating their utility for pseudo-label quality estimation in accent adaptation.

\subsection{Variable-budget Evaluation}\label{subsec:paraformer}
To provide a complementary, budget-sweep validation on a stronger ASR backbone, we report additional experiments using Paraformer on English test sets (L2-ARCTIC and IndicTTS English). These experiments isolate the effect of the post-selection module by applying it directly to a fixed candidate pool $\mathcal{C}$ without the FLMI pre-selection stage and analyzing WER under matched-hour budgets. We present three complementary views: (i) a WER--hours sweep to characterize budget-efficiency and compare multimodal filtering against text-only PPL, (ii) a random-sampling sanity check at matched hours, and (iii) a comparison to recent non-PPL selection baselines under matched-hour budgets. All-Ref serves as a supervised oracle, and All-Pseudo uses unfiltered pseudo-transcriptions.

\textit{Candidate pool differences from main experiments.} For IndicTTS, $\mathcal{C}$ is the same 30k pool from the main experiments, which was originally FLMI-preselected; thus, the Paraformer experiments do not re-run FLMI but apply post-selection directly on this fixed pool. For L2-ARCTIC, $\mathcal{C}$ is the training portion of the dataset (an in-domain pool), unlike the main experiments where we use VCTK as a cross-domain candidate pool. This difference means the Paraformer L2-ARCTIC setting tests in-domain pseudo-label purification, while the main L2-ARCTIC setting tests cross-domain selection under stronger shift.

\textbf{Purification-oriented variants (additional strategies).}
Besides Conf-P$^p$ (which selects utterances where perturbation yields improved hypotheses), we include two variants to disentangle whether gains come from perturbation-improvement signals or simply from selecting high-quality baseline pseudo-labels---similar to the filtering approach in prior work~\cite{rangappa2025efficient}.

\noindent (i) \textit{Stable-Base-P$^p$}: selects utterances whose \emph{baseline} hypothesis $H_0$ (without perturbation) is already high-quality, analogous to prior pseudo-label filtering methods that retain only high-confidence transcriptions. Unlike Conf-P, this strategy does not require improvement over the baseline; instead, it directly thresholds the baseline quality scores. Formally, an utterance $u$ is selected if:
\begin{equation}
  \hat{w}_0 \leq \tau_{\hat{w}}^{p}
  \;\land\;
  \bigl(
  c_0 \geq \tau_{c}^{100-p}
  \;\lor\;
  d_0 \leq \tau_{d}^{p}
  \bigr),
  \label{eq:stable-base}
\end{equation}
where $\hat{w}_0$, $c_0$, and $d_0$ are the predicted WER, cosine similarity, and Euclidean distance for the baseline hypothesis, and $\tau^p$ denotes the $p$-th percentile threshold computed from the \emph{baseline hypothesis distribution} over $\mathcal{P}$ (not the improvement distribution). For lower-is-better metrics ($\hat{w}$, $d$), we use the $p$-th percentile; for higher-is-better ($c$), we use the $(100{-}p)$-th percentile.

\noindent (ii) \textit{Conf-Stable-$p_1$-$p_2$}: inspired by curriculum learning, merges a Conf-P$^{p_1}$ subset (perturbation-improved hypotheses) with a Stable-Base-P$^{p_2}$ subset (baseline high-quality hypotheses):
\begin{equation}
  \mathcal{S}_{\text{conf-stable}}^{p_1, p_2} = \mathrm{Merge}\bigl(\mathcal{S}_{\text{conf}}^{p_1},\, \mathcal{S}_{\text{stable}}^{p_2}\bigr).
  \label{eq:conf-stable}
\end{equation}
For example, Conf-Stable-P70-P50 combines Conf-P70 and Stable-Base-P50. The operator $\mathrm{Merge}(\cdot)$ concatenates the two selected subsets; if an utterance appears in both, it is included twice, effectively upweighting it. This variant tests whether combining both selection paradigms---perturbation-improved and baseline-quality---yields additional benefit.

Results show that both strategies can be effective, suggesting that the multimodal signals themselves---not the perturbation mechanism alone---are the key drivers of quality estimation.

\begin{figure}[t]
  \centering
  \includegraphics[width=\columnwidth]{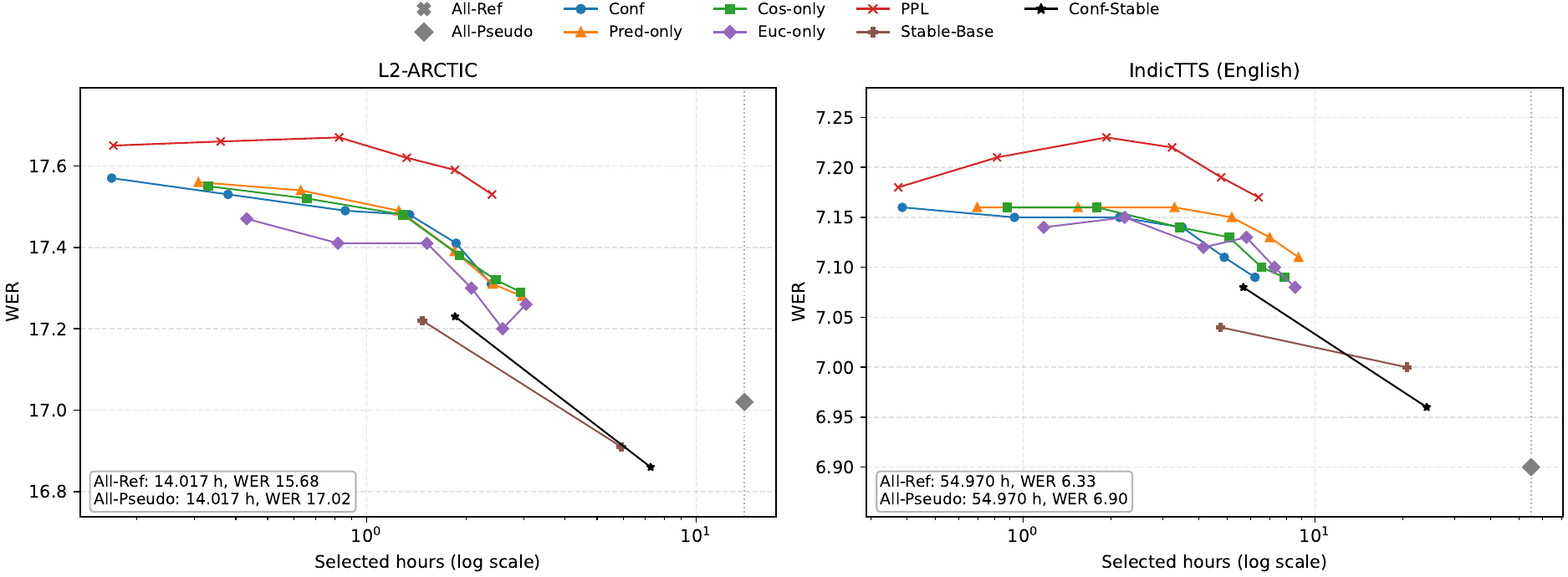}
  \caption{Paraformer post-selection: WER versus selected hours (log scale).}
  \label{fig:paraformer_efficiency}
\end{figure}

\textbf{Budget-sweep analysis (WER--hours trade-off).} Fig.~\ref{fig:paraformer_efficiency} summarizes full percentile sweeps as WER versus selected hours (log-scale). We extend the percentile range to $p \in \{50,60,70,80,90,95\}$ for finer-grained budget analysis. Three key observations emerge from the sweep:

\noindent (i) \textbf{Multimodal signals generally outperform text-only PPL}: Across both datasets, the PPL curve tends to lie above multimodal selection curves at comparable hour budgets. For example, at the $p{=}60$ operating point on L2-ARCTIC, Conf-P60 achieves 17.41\% WER at 1.867~h, compared to PPL-P60's 17.59\% at 1.851~h; on IndicTTS, Conf-P60 reaches 7.11\% at 4.884~h versus PPL-P60's 7.19\% at 4.766~h. This suggests that speech--text alignment signals capture quality information complementary to lexical perplexity.

\noindent (ii) \textbf{Data efficiency}: On L2-ARCTIC, the merged variant Conf-Stable-P70-P50 (Eq.~\ref{eq:conf-stable}) achieves 16.86\% WER at 7.272~h (51.9\% of All-Pseudo's 14.017~h), outperforming All-Pseudo's 17.02\%, suggesting data purification is possible in this setting. On IndicTTS, where All-Pseudo is already strong (6.90\% WER), post-selection subsets offer a trade-off between data budget and accuracy rather than strict WER improvement---e.g., Euc-Only-P50 reaches 7.08\% with only 15.5\% of the hours.

\noindent (iii) \textbf{Relative stability across signals}: Among multimodal variants, Euc-Only tends to perform competitively across hour budgets; the differences among Conf-P, Pred-Only, Cos-Only, and Euc-Only are generally small, indicating that all proposed multimodal signals tend to outperform text-only PPL and practitioners can choose based on computational convenience.

Because the mapping from percentile $p$ to selected hours depends on utterance-length and score distributions, we interpret these trends as directional observations rather than claiming a single globally optimal $p$.

\begin{figure}[t]
  \centering
  \includegraphics[width=\columnwidth]{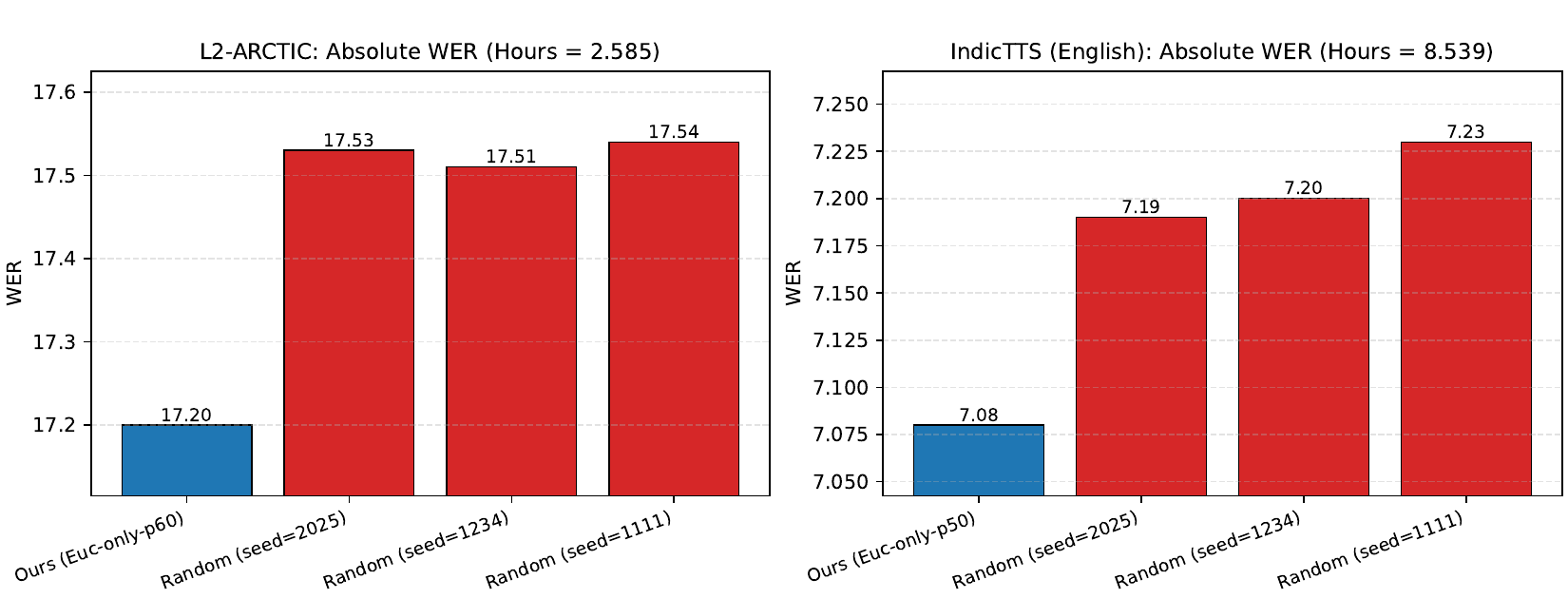}
  \caption{Performance comparison under matched audio-hour budgets: multimodal post-selection vs. random subsets (multiple seeds).}
  \label{fig:paraformer_random}
\end{figure}

\textbf{Sanity check (random baseline at matched hours).} Multimodal post-selection generally outperforms random sampling at matched durations, demonstrating its ability to avoid low-quality pseudo-labels. Fig.~\ref{fig:paraformer_random} compares our method to multiple random controls (different random seeds) under representative budgets of roughly 15--20\% of the full pseudo-labeled hours (All-Pseudo). We select the operating point whose hours most closely match the target budget from our percentile sweep: Euc-Only-P60 on L2-ARCTIC (17.20 WER at 2.585~h) and Euc-Only-P50 on IndicTTS (7.08 WER at 8.539~h). On L2-ARCTIC, the selected subset outperforms three random controls at the same duration (17.51--17.54 WER). On IndicTTS, the selected subset similarly improves over random subsets matched by hours (7.19--7.23 WER). These comparisons reduce the chance that improvements arise from accidental subsampling.

\textbf{Comparison to recent non-PPL selection baselines.} Our quality-based selection is competitive with model-agreement or SSL-classification baselines under matched-hour budgets. We compare to two baselines from~\cite{rangappa2025efficient}: (i) CER-Consistency, which selects utterances based on the average CER consistency across three ASR hypotheses (Paraformer, Zipformer, and Parakeet), keeping segments with $\mathrm{CER}_{\mathrm{avg}}(s) < \tau$ (e.g., 5\%) where $\mathrm{CER}_{\mathrm{avg}}(s) = \frac{1}{3}\left(\mathrm{CER}_{\mathrm{p-z}}(s)+\mathrm{CER}_{\mathrm{p-k}}(s)+\mathrm{CER}_{\mathrm{z-k}}(s)\right)$, and (ii) SSL-WER-Classifier, which uses an SVM on SSL features to predict WER and filter low-quality pseudo-labels (below 0.5 as high-quality; above 0.5 as low-quality). Since these baselines filter based on baseline hypothesis quality rather than perturbation-improvement, we use Stable-Base-P50 as a representative of our approach for a fair comparison. Under the same hour budget on L2-ARCTIC (Table~\ref{tab:paraformer_sota_compare}), Stable-Base-P50 outperforms both baselines across multiple random seeds, suggesting that multimodal quality signals can provide competitive or better selection than model-agreement or SSL-feature classification.

\begin{table}[t]
  \centering
  \caption{Comparison to recent selection baselines under matched hours (L2-ARCTIC, Paraformer).}
  \label{tab:paraformer_sota_compare}
  \begin{tabular}{l c c}
    \hline
    \textbf{Method} & \textbf{Hour} & \textbf{WER} \\
    \hline
    Stable-Base-P50 (ours) & 5.923 & \textbf{16.91} \\
    CER-Consistency (seed 2025) & 5.923 & 16.99 \\
    CER-Consistency (seed 1234) & 5.923 & 17.06 \\
    CER-Consistency (seed 1111) & 5.923 & 17.02 \\
    SSL-WER-Classifier (seed 2025) & 5.923 & 17.22 \\
    SSL-WER-Classifier (seed 1234) & 5.923 & 17.21 \\
    SSL-WER-Classifier (seed 1111) & 5.923 & 17.23 \\
    \hline
  \end{tabular}
\end{table}

\section{Conclusions}
Our experiments show that multimodal consistency signals---speech--text alignment in SONAR space and predicted WER---provide a practical reference-free criterion to filter pseudo-labels for ASR accent adaptation. Our main finding is that these multimodal signals provide more reliable quality estimation than text-only perplexity across different backbones and domain conditions. In an in-domain setting, \(\sim\)1.5k selected utterances achieve 10.91\% WER, approaching the 10.45\% from 30k supervised labels. In a cross-domain setting with a mismatched candidate pool, consistency filtering avoids degradation that unfiltered pseudo-labels cause. Matched-hour experiments on a stronger ASR backbone confirm that the post-selection module generalizes across backbones and outperforms random sampling and recent selection baselines under matched budgets. Future work includes inductive extensions, adaptive thresholds, and continual adaptation.
\bibliographystyle{IEEEtran}
\bibliography{sample}

\end{document}